\documentclass[letterpaper, 10 pt, conference]{ieeeconf}  % Comment this line out if you need a4paper

\IEEEoverridecommandlockouts                              % This command is only needed if 
\usepackage{graphics} % for pdf, bitmapped graphics files
\usepackage{amsmath} % assumes amsmath package installed
\usepackage{amssymb}  % assumes amsmath package installed
\usepackage{bm}
\usepackage[exponent-product = \cdot]{siunitx}
\usepackage{xcolor}
\usepackage{algorithm} %for Pseudocode
\usepackage{booktabs} % for Tables 
\usepackage[mode=buildnew]{standalone}
\usepackage{comment}

\newtheorem{problem}{Problem}

\newcommand{\sysX}{\bm x}
\newcommand{\sysU}{\bm u}
\newcommand{\sysY}{\bm y}
\newcommand{\pathParam}{s}
\DeclareMathOperator*{\argmax}{arg\,max}

\title{\LARGE \bf
Multi-Objective Optimization of a Path-following MPC for Vehicle Guidance: A Bayesian Optimization Approach
}

\author{Ali Gharib$^{2}$, David Stenger$^{2}$, Robert Ritschel$^{1}$, and Rick Voßwinkel$^1$ %% <-this % stops a space
\thanks{$^{1}$Development Center Chemnitz/Stollberg, IAV GmbH, Stollberg, Germany  {\tt\small \{Robert.Ritschel, Rick.Vosswinkel\}@iav.de}}%
\thanks{$^{2}$Institute of Automatic Control, RWTH Aachen University, Aachen, Germany
        {\tt\small Ali.Gharib@rwth-aachen.de, D.Stenger@irt.rwth-aachen.de}}%
}

\begin{document}
	\begin{comment}
\IEEEoverridecommandlockouts
\IEEEpubid{{“\copyright~2021 IEEE.  Personal use of this material is permitted.  Permission from IEEE must be obtained for all other uses, in any current or future media, including reprinting/republishing this material for advertising or promotional purposes, creating new collective works, for resale or redistribution to servers or lists, or reuse of any copyrighted component of this work in other works.\hfill} \hspace{\columnsep}} 
\end{comment}

%Copyright notice
%\pagestyle{empty}
\twocolumn[
\begin{@twocolumnfalse}
	
	%	Author’s pre-print (ie pre-refereeing) /post-print (ie final draft post-refereeing) accepted for publication/published in the IEEE Transactions on XXX (Jan. 2021).\\
	
	%This work has been accepted for publication at 2021 European Control Conference.\\
	
	\copyright 2021 IEEE.  Personal use of this material is permitted.  Permission from IEEE must be obtained for all other uses, in any current or future media, including reprinting/republishing this material for advertising or promotional purposes, creating new collective works, for resale or redistribution to servers or lists, or reuse of any copyrighted component of this work in other works.\\
	%DOI: \url{}\\ %fill this for post-print
	%URL: \url{}\\ %fill this for post-print
	
	%	Cite: J. Doe, A. Einstein, “Title of the article,” \textit{IEEE Transactions on xxx}, vol. xx, no. xx, id. xx, Jan. 2021. \\ %fill this for post-print
	
\end{@twocolumnfalse}
]

\maketitle
\thispagestyle{empty}
\pagestyle{empty}

%%%%%%%%%%%%%%%%%%%%%%%%%%%%%%%%%%%%%%%%%%%%%%%%%%%%%%%%%%%%%%%%%%%%%%%%%%%%%%%%
\begin{abstract}
This paper tackles the multi-objective optimization of the cost functional of a path-following model predictive control for vehicle longitudinal and lateral control. While the inherent optimal character of the model predictive control and the direct consideration of constraints gives a very powerful tool for many applications, is the determination of an appropriate cost functional a non-trivial task. This results on the one hand from the number of degrees of freedom or the multitude of adjustable parameters and on the other hand from the coupling of these. To overcome this situation a Bayesian optimization procedure is present, which gives the possibility to determine optimal cost functional parameters for a given desire. Moreover, a Pareto-front for a whole set of possible configurations can be computed.
\end{abstract}

%%%%%%%%%%%%%%%%%%%%%%%%%%%%%%%%%%%%%%%%%%%%%%%%%%%%%%%%%%%%%%%%%%%%%%%%%%%%%%%%
\section{INTRODUCTION}
\begin{figure}
	\centering
	%\scriptsize
	\includegraphics[width=1.0\linewidth]{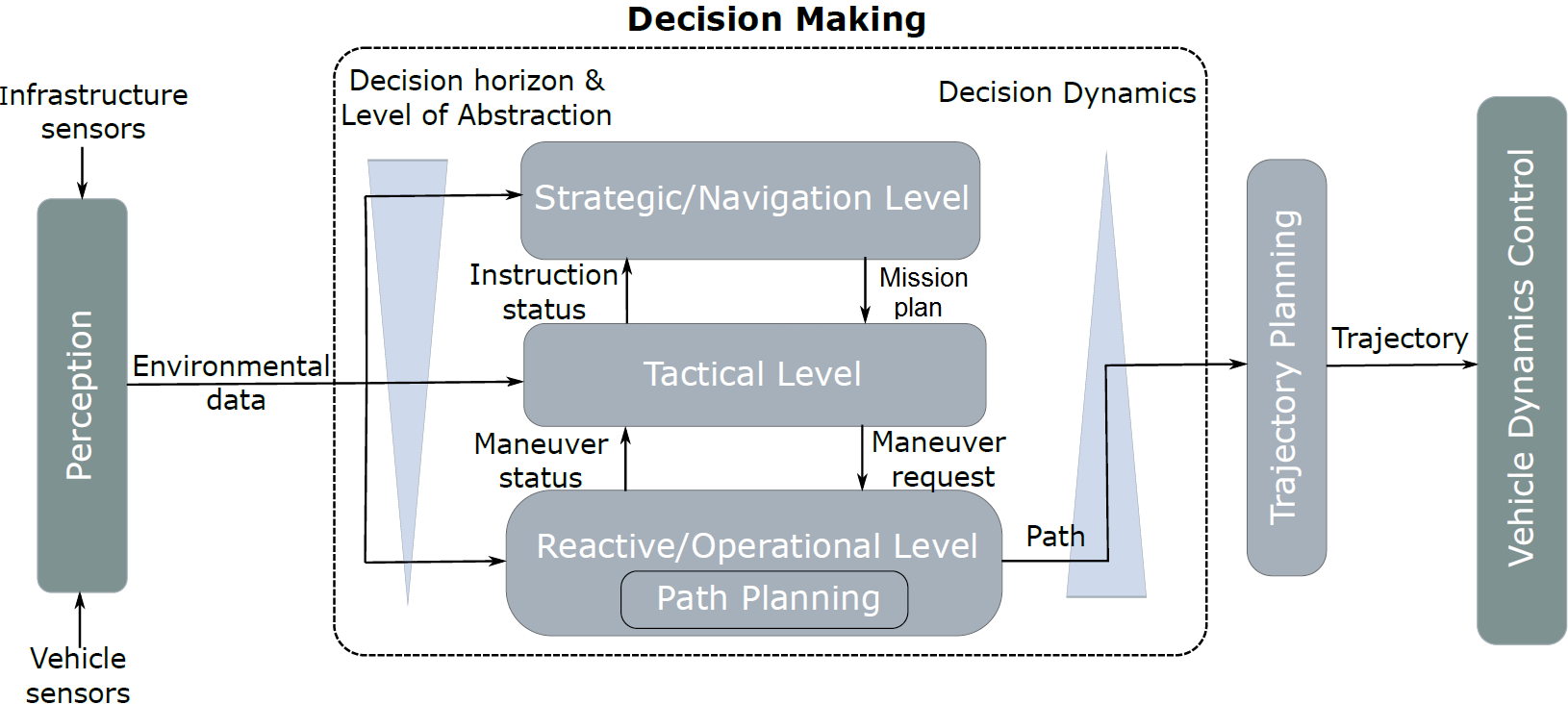}
	\caption{Automation Structure, c.f. \cite{schroedel2019}}
	\label{fig:AutomationStructure}
\end{figure}

In the last decade, we see a continuously increasing activity in the research and development of automated vehicles.  The interests of academia and industry go from small robotic solutions to people mover to passenger cars or even trucks. These systems mainly consist of the parts, perception, decision making, and control, which are generally split into sub-functionalities, see Fig.~\ref{fig:AutomationStructure}. The perception collects all sensor data and fuses them to create a data basis for the subsequent steps. Thereafter is the decision-making with a hierarchic structure of planning layers, which more and more precise the requested actions. These actions are forwarded via a control structure to the actuators.

While the whole automation structure~\cite{schroedel2019}, see Fig.~\ref{fig:AutomationStructure}, is significant for the overall performance the focus of the paper is the operational planning and the control of the vehicle guidance. A key discipline for motion planning and control is mathematical optimization, e.\,g. \cite{Betts1998} since a safe and comfortable driving behavior could not trivially be realized without any (in most cases numerical) optimization.
A well-known and established method for the proposed issue is model predictive control (MPC) \cite{GARCIA1989}. Which gives based on a cost functional and a dynamical model of the plant an optimal control input. 
Although the calculated input signal is optimal with respect to the established cost functional and the system model, the formulation and parametrization of a suitable cost functional itself is a complicated problem and can also be seen as an optimization problem. Customer-relevant requirements such as driving comfort cannot explicitly be considered in the MPC formulation either due to insufficient control horizons or structural restrictions. Furthermore, multiple requirements can be conflicting, therefore a trade-off needs to be found. Solving these problems manually can be tedious and possibly suboptimal. This paper will illustrate a procedure to solve this problem using mathematical optimization..%and find optimal cost functionals. 

%In the case without additional constrains, this %optimization problem can be written as: 

%\begin{equation} \label{eq:optproblemIntro} %\boldsymbol{\theta}_c^{\text{opt}} = 
%\min_{\boldsymbol{\theta}_c} %(C_1(\boldsymbol{\theta}_c),C_2(\boldsymbol{\theta}_c),..%.,C_n(\boldsymbol{\theta}_c)), %+ \epsilon
%\end{equation}
%where $C_i(\boldsymbol{\theta}_c)$ are possibly conflicting nonlinear cost functions concerning the whole driving cycle. Controller parameters, ${\theta}_c$, are unknown initially. In the case of MPC these can for example include elements of the weight matrices within the objective function.

The performance function(s) can either be evaluated in an experiment or simulation of the closed-loop. Therefore, the problem is a black-box optimization problem, where a closed analytical form of the expensive-to-evaluate objective functions is not given. As a result, gradients are not available and convexity cannot be guaranteed. Therefore, optimization algorithms relying on a high number of function evaluations such as genetic algorithms or numerically approximated gradients may not be the most efficient.

An additional challenge is presented by dealing with conflicting goals in optimization. By using prior preference information, either all but one objective can be converted into constraints resulting in a constrained optimization problem or all objectives can be integrated via weighting in one objective. Here we also consider the case without prior preference information and focus on searching for the set of Pareto-optimal MPC parametrizations (the set of non-dominated compromises). This set can be used to visualize the trade-off between the conflicting goals and serve as a decision support system.   

Recently, Bayesian optimization is becoming increasingly popular for the tuning of model predictive controllers. Andersson et al. \cite{Andersson.2016} used constrained Bayesian optimization to search for an optimal soft constraint parametrization for stochastic collision avoidance. In \cite{Piga.2019} model parameters and MPC parameters are simultaneously considered during optimization whereas \cite{Lucchini.2020} minimize the yaw rate difference in torque vectoring for high-performance electric vehicles. Insurance of satisfactory worst-case behavior with respect to distributed model plant miss-matches can be achieved while minimizing expected integral tracking error using Bayesian optimization \cite{stenger2020robust}. However, to the best of the authors' knowledge previous work in MPC tuning with Bayesian optimization has not considered Pareto optimization.

Therefore contributions of this paper are to
\begin{itemize}
    \item apply Bayesian optimization to model predictive path-following control for autonomous driving
    \item and perform Pareto optimization to make the trade-off between different objectives transparent.
\end{itemize}

%\textcolor{red}{Mehrzieloptimierung von MPC mit anderen Optimierern?, MPC Parameter tuning für Trajektoriefolgeregelung im Kontext Autonomatisiertes fahren?}

For that purpose the paper is organized as follows:
After this introduction, a brief summary of the used MPC formulation is given. In Section~\ref{sec:method} we will present the optimization technique for the cost function, followed by illustrative results and a discussion on the achieved quality. The paper ends with some concluding remarks and an outlook to further research.

\section{Background and Problem Formulation}
In this section, the used MPC formulation, the simulation framework as well as the weighting parameter optimization problem are introduced.

\subsection{Model Predictive Path-Following Control}
The proposed approach for optimization of weighting parameters is demonstrated using the MPC realization of \cite{Ritschel2019}. This MPC is designed for longitudinal and lateral vehicle guidance in the context of highly automated driving. The implementation is based on a special nonlinear MPC approach called model predictive path-following control (MPFC) as proposed in \cite{Faulwasser.2013}. The advantage of this MPC approach is that the assignment of the reference trajectory and the computation of inputs to track this trajectory is handled simultaneously at the run-time of the controller.
We recall the path-following problem and the path-following MPC formulation from \cite{Ritschel2019} in the following.

%%%%%%%%%%%%%%%%%%%%%%%%%%%%%%%%%%%%%%%%%%%%%%%%%%%%%%%%%%%%%%%%%%%%%%%%%%%%
% Model
%%%%%%%%%%%%%%%%%%%%%%%%%%%%%%%%%%%%%%%%%%%%%%%%%%%%%%%%%%%%%%%%%%%%%%%%%%%%
To describes the longitudinal and lateral dynamics of the vehicle, a continuous-time nonlinear system in the form
\begin{subequations}
\label{eq:system}
\begin{align}
	\dot{\sysX}(t) &= \bm f(\sysX(t),\sysU(t)), \quad \sysX(t_0)=\sysX_0\\
	\bm \sysY(t) & = \bm h(\sysX(t)) \label{eq:systemOutput}
\end{align}
\end{subequations}
with the maps $\bm f: \mathbb{R}^{9} \times \mathbb{R}^{2} \rightarrow \mathbb{R}^{9}$ and $\bm h: \mathbb{R}^{9} \rightarrow \mathbb{R}^{3}$ is used. 
The state vector $\sysX =[x,\, y,\, \psi,\, \dot{\psi},\, \beta,\, v,\, v_{ref},\, \delta_{s},\, \delta_{s,ref}]^\top$ consists of the vehicle states, i.e.\ the coordinates of the center of gravity in an inertial frame $x$ and $y$, the yaw angle $\psi$, the yaw rate $\dot{\psi}$, the side slip angle $\beta$, the velocity $v$, the target velocity $v_{ref}$, the steering wheel angle $\delta_{s}$ and the steering wheel target angle $\delta_{s,ref}$.
The control input $\sysU=[a_{ref},\, \omega_{s,ref}]^\top$ contains the target acceleration $a_{ref}$ and the steering wheel target angular velocity $\omega_{s,ref}$. In addition, the output vector $\sysY=[x_f,\, y_f,\, \psi]^\top$ includes the coordinates of the middle of the front axle $x_f$ and $y_f$ as well as the vehicle orientation $\psi$. The closed sets of state and input constraints are given by $\sysX\in\mathcal{X}\subseteq\mathbb{R}^{9}$ and $\sysU\in\mathcal{U}\subseteq\mathbb{R}^{2}$. For a more detailed description of the vehicle model we refer to \cite{Ritschel2019}.

%%%%%%%%%%%%%%%%%%%%%%%%%%%%%%%%%%%%%%%%%%%%%%%%%%%%%%%%%%%%%%%%%%%%%%%%%%%%
% path-following problem
%%%%%%%%%%%%%%%%%%%%%%%%%%%%%%%%%%%%%%%%%%%%%%%%%%%%%%%%%%%%%%%%%%%%%%%%%%%%
The primary objective is that the system \eqref{eq:system}, which describes the vehicle behavior, follows a given geometric reference in the output space \eqref{eq:systemOutput}. This geometric reference is called path $\mathcal{P}$. We represent it by a continuously differentiable curve
\begin{align}
	\label{eq:path}
    \mathcal{P} = \left\{\bm p(\pathParam) \in\mathbb{R}^{3}\, |\, \pathParam \in [0,\pathParam_{\mathrm{max}}] \mapsto \bm p(\pathParam)\right\},
\end{align}
where, the variable $\pathParam \in \mathbb{R}$ is called path parameter and $\bm p(\pathParam)$ is a parameterization of $\mathcal{P}$.
For the vehicle guidance task $\mathcal{P}$ is represented with $\bm p(\pathParam) =[x_{ref}(\pathParam),\, y_{ref}(\pathParam),\, \psi_{ref}(\pathParam)]^\top$, where $x_{ref}$, $y_{ref}$ and $\psi_{ref}$ describe the coordinates of the center line of the driving lane and the reference orientation for the vehicle as a function of the path parameter $\pathParam$. Note that in our case $\pathParam$ is equivalent to the distance travelled along the path.
The conceptual idea of MPFC is, that the controller determines the input $\sysU(t)$ to converge to reference path $\mathcal{P}$ as well as the time evolution $\pathParam(t)$ simultaneously at the run-time. With the deviation from the path
\begin{align}
	\label{eq:PathDeviation}
	\bm e(t) &:= \bm h(\sysX(t)) - \bm p(\pathParam(t)), 
\end{align}% \todo{Ein Bild zur Verdeutlichung?}
this output path-following problem can be formulated as follows: 
\begin{problem}
Given the system \eqref{eq:system} and the geometric reference path $\mathcal{P}$ \eqref{eq:path}, design a controller that computes $\sysU(t)$ and $\pathParam(t)$ and guarantees:
\begin{enumerate}
	\item Path convergence: The system output $\sysY$ converges to the path $\mathcal{P}$ such that $\lim\limits_{t \rightarrow \infty}{\left\|\bm e(t)\right\| = 0}$.
  \item Velocity convergence: The path velocity $\dot{\pathParam}(t)$ converges to a predefined evolution $\dot{\pathParam}_{ref}(t) \geq 0$ such that $\lim\limits_{t \rightarrow \infty}{\left\|\dot{\pathParam}(t) - \dot{\pathParam}_{ref}(t)\right\| = 0}$.
  \item Constraint satisfaction: The state and input constraints $\mathcal{X}$ and $\mathcal{U}$ are satisfied $\forall t \in [t_0, \infty)$. 
\end{enumerate}
\end{problem}

In order to solve the path-following problem using a MPC approach, the path parameter $\pathParam$ is treated as a virtual state whose time evolution is described by a differential equation termed \emph{timing law}. In our case the timing law is defined as a single integrator
\begin{align}
	\label{eq:TimeLaw}
	\dot{\pathParam}(t) &:= \vartheta(t),
\end{align}
where $\vartheta \in \mathcal{V} \subset \mathbb{R}$ is an additional (virtual) control input of the MPC. 

%%%%%%%%%%%%%%%%%%%%%%%%%%%%%%%%%%%%%%%%%%%%%%%%%%%%%%%%%%%%%%%%%%%%%%%%%%%%
% OCP
%%%%%%%%%%%%%%%%%%%%%%%%%%%%%%%%%%%%%%%%%%%%%%%%%%%%%%%%%%%%%%%%%%%%%%%%%%%%
To solve the path-following problem a sampled-data MPC strategy is used. Predicted system states and inputs are denoted by $\bar{\sysX}(\cdot)$ and $\bar{\sysU}(\cdot)$. The cost functional to be minimized for the prediction horizon $T_p$ is given by
\begin{align}
	\label{eq:MPCCostFunction}
	J&(\sysX(t_k),\,\pathParam(t_k),\bar{\sysU}(\cdot),\bar{\vartheta}(\cdot))
	=\\
	&\int_{t_k}^{t_k+T_p}\! \left\|
		\begin{matrix}
	\bar{\bm e}\\
	a_{lat}(\bar{\sysX})\\
	\end{matrix}\right\|^2_Q +  \left\|
		\begin{matrix}
	\bar{\sysU} \\
	\bar{\vartheta} - \vartheta_{ref}\\
	\end{matrix}\right\|^2_R	\, \mathrm{d}\tau 
			+ \left\|
			\begin{matrix}
	\bar{\bm e}\\
	a_{lat}(\bar{\sysX})\\
	\end{matrix}\right\|^2_P \notag
\end{align}
where $Q = \operatorname{diag}(q_x,q_y,q_\psi,q_{a})$, $P = \operatorname{diag}(p_x,p_y,p_\psi,p_{a})$ and $R = \operatorname{diag}(r_a,r_\omega,r_\vartheta)$ are positive definite weighting matrices. Furthermore, $a_{lat}(\sysX) = v(t)\, \dot{\psi}(t)$ is an approximation of the lateral acceleration of the vehicle. Note that the last term of \eqref{eq:MPCCostFunction} represents the terminal cost.

The optimal control problem (OCP) solved at every discrete sampling time instance $t_k = k\,T_s$ in a receding horizon fashion reads
\begin{subequations}
	\label{eq:OCP}
	\begin{align}
		\min_{\bar{\sysU}(\cdot), \bar{\vartheta}(\cdot)} &J(\sysX(t_k),\pathParam(t_k),\bar{\sysU}(\cdot),\bar{\vartheta}(\cdot))
		\intertext{subject to the constraints for all $\tau \in [t_k,t_k+T_p]$}
		\dot{\bar{\sysX}}(\tau) &= \bm f(\bar{\sysX}(\tau),\bar{\sysU}(\tau)), \quad \bar{\sysX}(t_k)=\sysX(t_k) \label{eq:OCP_Sys}\\
		\dot{\bar{\pathParam}}(\tau) &= \bar{\vartheta}(\tau), \quad \bar{\pathParam}(t_k)=\pathParam(t_k)\label{eq:OCP_TimeLaw}\\
		\bar{\bm e}(\tau) &= \bm h(\bar{\sysX}(\tau)) - \bm p(\bar{\pathParam}(\tau))\label{eq:OCP_PathDeviation}\\
		\bar{\sysU}(\tau) &\in \mathcal{U}, \quad	\bar{\sysX}(\tau) \in \mathcal{X} \label{eq:OCP_StateInputConstraints}\\
		\bar{\pathParam}(\tau) &\in [0,\pathParam_{max}], \quad \bar{\vartheta}(\tau) \in \mathcal{V} \label{eq:OCP_VirtualStateInputConstraints}\\
		\bm h_c(\bar{\sysX}(\tau),\bar{\sysU}(\tau)) &\leq \bm 0. \label{eq:OCP_ComplexConstraints}
	\end{align}
\end{subequations}
Here, the dynamics of system \eqref{eq:system} and the time law \eqref{eq:TimeLaw} with their respective initial conditions are stated by the constraints \eqref{eq:OCP_Sys} and \eqref{eq:OCP_TimeLaw}. The the deviation of the system output from the path \eqref{eq:PathDeviation} is represented by \eqref{eq:OCP_PathDeviation}. The state and input constraints are enforced by \eqref{eq:OCP_StateInputConstraints} and \eqref{eq:OCP_VirtualStateInputConstraints}. Additionally, \eqref{eq:OCP_ComplexConstraints} defines further constraints with the constraint function $\bm h_c$, see \cite{Ritschel2019}. The solution of \eqref{eq:OCP} results in the optimal input trajectory $\bar{\sysU}^\star_k(\cdot)$.  Finally, this optimal input $\bar{\sysU}^\star_k(\cdot)$ is applied to the system \eqref{eq:system}
\begin{align}
	\forall t \in [t_k, t_k + T_s): \sysU(t) = \bar{\sysU}_k^\star(t).
\end{align}
At $t_k + T_s$, the OCP \eqref{eq:OCP} is solved again using new initial conditions.

%%%%%%%%%%%%%%%%%%%%%%%%%%%%%%%%%%%%%%%%%%%%%%%%%%%%%%%%%%%%%%%%%%%%%%%%%%%%
% Simulation Framework
%%%%%%%%%%%%%%%%%%%%%%%%%%%%%%%%%%%%%%%%%%%%%%%%%%%%%%%%%%%%%%%%%%%%%%%%%%%%
\subsection{Simulation Framework}

\begin{figure}[b]
	\centering
	\scriptsize
	\includegraphics[width=1.0\linewidth]{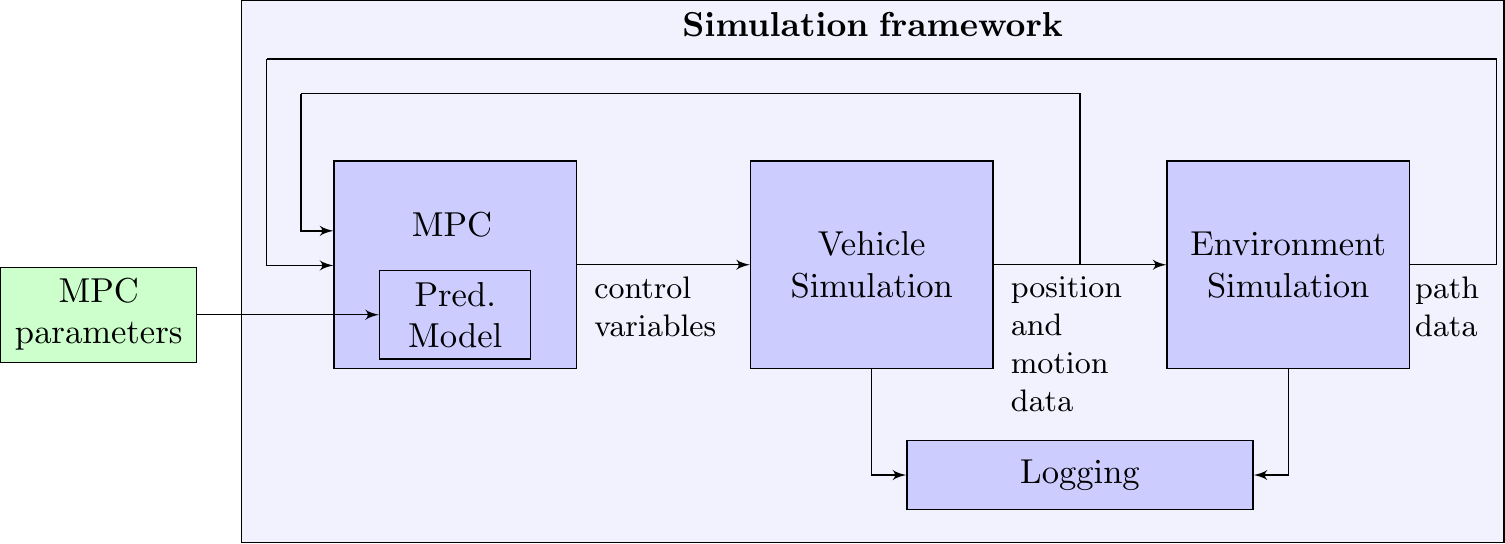}
	\caption{Closed-loop Simulation}
	\label{fig:SimulationFramework}
\end{figure}

In order to find optimal parameter settings for the weighting matrices of the MPC, their performance has to be evaluated for different parameter sets. For this purpose, we use a closed-loop simulation as shown in Fig.~\ref{fig:SimulationFramework}, which consists of the MPC, a plant, and a data logging function.

\begin{figure}
	\centering
	\scriptsize
	\includegraphics[width=1.0\linewidth]{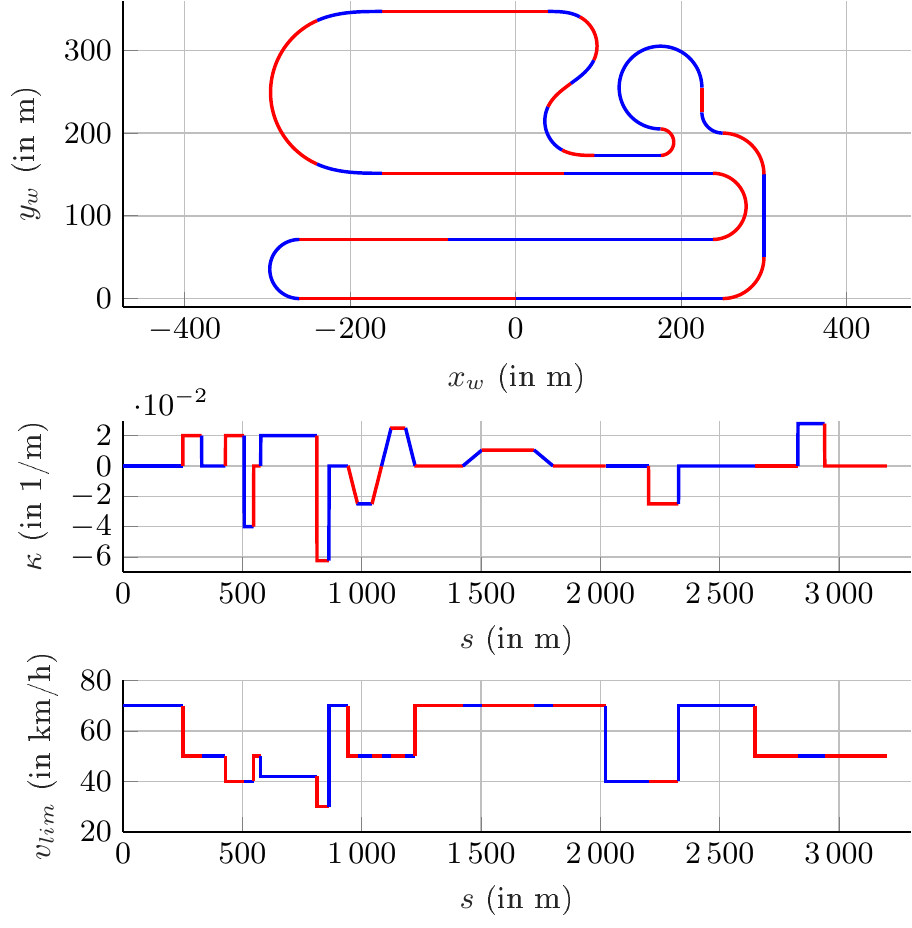}
	\caption{Bird’s eye view, curvature $\kappa$ and speed limit $v_{lim}$ of the virtual road loop used for performance evaluation. The different sections of the route are visualized by the colors red and blue, which is only used for clarity.}
	\label{fig:RouteModel}
\end{figure}

The MPC uses the outputs of the simulated plant to determine optimal control variables. Furthermore, it has an interface that is used to specify the current parameter set. The plant consists of two parts: a dynamic vehicle model and an environment model. The vehicle model represents the dynamic lateral and longitudinal behavior of the ego vehicle in dependence on the control variables. Its outputs are the world coordinates, the orientation, and the states of the ego vehicle, such as the velocity. Based on this, the environment model provides input data for the MPC, which are used to determine the parameterization \eqref{eq:path} of the path $\mathcal{P}$ and the target path velocity $\dot{\pathParam}_{ref} = \vartheta_{ref}$. Therefore, the environment model simulates an idealized lane detection using a predefined virtual two-way road with two lanes. In this article, we use a road loop as shown in Fig.~\ref{fig:RouteModel} representing an urban-like scenario with curves of small radius.

For performance evaluation of the MPC parameterization, the data for a complete drive of the road loop starting at $x_w=0$ and $y_w=0$ is used. The simulation data is stored by a data logging function. Please note that in our experiments the vehicle model of the simulation and the prediction model of the MPC were identical. However, this is not a prerequisite for our MPC tuning method proposed in the following. It works even if there are model deviations because it is formulated as a black-box optimization.  

\begin{comment}
\begin{figure}
\centering
\includestandalone[width=1.0\linewidth]{texFigures/BirdviewPlot2}
\caption{Bird’s eye view, curvature $\kappa$ and speed limit $v_{lim}$ of the virtual road loop used for performance evaluation. The different sections of the route are visualized by the colors red and blue, which is only used for clarity.}
\label{fig:RouteModel}
\end{figure}
\end{comment}

\subsection{Parameter Tuning Problem}
Different situations, like urban drive, highway driving, or parking require different controlling behavior to ensure the desired ride quality. 
% warum modellieren wir Komfort und warum Dynamik? 
Since comfort is probably the most important wish of passengers during autonomous driving, e.g. during longer journeys, the following section examines how driving behavior can be evaluated. 
Later, the evaluation of driving behavior with regard to dynamics is discussed, since the responsiveness of the vehicle to changes in speed can suffer from the desired high level of comfort. Especially in urban driving scenarios, a dynamic driving style may be explicitly desired in order not to obstruct the traffic flow. 

%Besides the structure of the MPC, there are many parameters that can be adjusted to modify the car's driving behavior. Namely the entries of the weighting matrices $Q$, $R$ and $P$ of the MPC cost function \eqref{eq:MPCCostFunction} are considered in the optimization. The Parameters x and y are set equal to reduce the optimization’s dimension.

%The aim is to get the best possible parametrizations, by maximizing comfort, dynamic, and/or path following. The workflow of the optimization and the inputs and outputs of the simulation-framework is shown in {\color{red}figure X}. 

\textbf{Comfort}:
Comfort is defined as the subjective well-being of the passengers during a ride \cite{Fahrwerkhandbuch_Ersoy2017}. Comfort can be divided into the passenger’s psychological and physiological well-being, with many factors that are hard to measure \cite{Wang_ComfortMeasuring}. The simulation framework only simulates the physiological effects. Therefore psychological effects won't be considered in the optimization.
%In addition to that the simulation framework described above, the vehicle is modeled as a point mass and the track is only varying in horizontal (x and y) directions. This work will concentrate on the factors, which have a physiological impact on the passenger’s body. 
%(The acceleration signal is used for the calculation of various characteristic values described in ISO 2631 {\color{red}(Quelle)}. The characteristic values are frequency-weighted accelerations.)
%Even unique events can have a significant impact on the human’s comfort or well-being \cite{BELLEM201645}. 
According to \cite{strandemar2005objective} passengers are more sensitive to high jerk, which is the first time derivative of acceleration. Regardless of the jerk, high accelerations are perceived as uncomfortable, especially at high velocities.
To quantify the comfort during a complete drive of the road loop we use the mean-squared-error of the jerk $j$ 
\begin{align} 
    \Bar{E}_{jerk} = \frac{1}{N_k}\sum_{k=1}^{N_k}(j(t_k))^2
   	\label{eq:errorJerk}
\end{align}
were $N_k$ is the number of simulation time steps needed for a complete drive. Note that we assume a jerk of zero as optimal. Also high longitudinal acceleration should be avoided and limited to $a_{x_{min}} = \SI{-3.5}{\meter\per \square\second}$ and $a_{x_{max}} = \SI{2.5}{\meter\per \square\second}$ (see \eqref{eq:FOP_cons1}),  like it is done in common adaptive cruise control systems \cite{Fahrerassistenz_Handbuch2015}. %{\color{red}(ist auch schon in der MPC als constraint gesetzt?)}.

In addition to jerk and acceleration, the study \cite{KomfortorientierteRegelung} shows that deviation from the lane centerline makes passengers feel less comfortable.
The lane tracking performance for driving the road loop shown in Fig.~\ref{fig:RouteModel} is modeled with the mean
\begin{align} 
    \Bar{E}_{lat} =& \frac{1}{N_k}\sum_{k=1}^{N_k}(e_{lat}(t_k))^2
    %\\e_{lat}(t_k) =& \sin(\psi(t_k))(x_{ref}-x_f(t_k))\\ &+ \cos(\psi(t_k))(y_{ref}-y_f(t_k))
   	\label{eq:errorLat}
\end{align}
of the squared lateral deviations $e_{lat}$ from the lane centerline to the middle of the front axle.
To ensure the vehicle will not leave the road we allow the vehicle to leave the center of the lane until the center of the vehicle crosses the lane boundaries.% {\color{red}(komischer Ausdruck? Alternattiv: lateral deviation is limited to half lane width)} (see \eqref{eq:FOP_cons2}).

\textbf{Dynamic}:
Dynamical driving can be realized by driving as fast as possible without exceeding speed limits. 

To describe the dynamics of a drive, the mean velocity error
%Mean squared error & 
\begin{align} 
   \Bar{E}_{v} =  \frac{1}{N_k}\sum_{k=1}^{N_k}(e_v(t_k))^2, \quad e_v(t_k) = v_{lim}(t_k)-v(t_k)
   \label{eq:errorV}
\end{align}
represented by the mean-squared error of the speed limit $v_{lim}$ from the actual ego velocity $v$ is used. 
Like the characteristic values $\Bar{E}_{lat}$ and $\Bar{E}_{jerk}$, $\Bar{E}_{v}$ is also calculated based on the simulation data for a complete drive of the road loop.

%% Was wird genau optimiert und welche Constraints werden gesetzt. 

Besides the structure of the MPC, there are many parameters that can be adjusted to modify the driving behavior of the vehicle. In this paper, we optimize the entries of the weighting matrices $Q$, $R$ and $P$ of the MPC cost function \eqref{eq:MPCCostFunction}. Thereby the parameters in $Q$ and $P$ are set equal in order to reduce the number of optimization variables. The vector $\bm m = [q_x,\, q_y,\, q_\phi,\, q_a,\, r_a,\, r_\omega,\, r_\vartheta]$ represents the parameters, that will be optimized in the optimization problem
%{\color{red}(soll hier noch erwähnt werden, dass wir eine gute hand-tuned Parametrisierung haben?)}
%\begin{subequations}\label{eq:FullOptProblem}
%	\begin{align} 
%	\min_{\boldsymbol{\theta}_c} \quad & C_{\text{EITE}}(\boldsymbol{\theta}_c)  \label{eq:FOP}\\[7pt]%\]  % _{\theta_e \sim \mathcal{N}}
%	s.\,t.\quad \, \, & \boldsymbol{\theta}_c \in \mathbb{Z}^3 \times \mathbb{R}^{3} \label{eq:2b} \\[7pt]
%	& \boldsymbol{\theta}_c = f(\Bar{E}_{jerk}, \Bar{E}_{lat}, \Bar{E}_{v}) \label{eq:2c} \\[7pt]
%	& a_{x_{min}} \leq  ({a_x})_{1,...,N_k} \leq a_{x_{max}}  \label{eq:FOP_cons1} \\[7pt]
%	& \vert (e_{lat})_{1,...,N_k}\vert \leq  s/2  \label{eq:FOP_cons2}  
%	\end{align}
%\end{subequations}	
%\begin{subequations}\label{eq:FullOptProblem}
%	\begin{align} 
%	\min_{\bm m} \quad & [\Bar{E}_{lat}(\bm m), \Bar{E}_{jerk}(\bm m), \Bar{E}_{v}(\bm m)] \label{eq:FOP}\\[7pt]%\]  % _{\theta_e \sim \mathcal{N}}
%	s.\,t.\quad \, \, 
%	& \bm m \in\, \bm M  \notag\\[7pt]
%	& a_{x_{min}} \leq  {a_x}(t_k) \leq a_{x_{max}}  \label{eq:FOP_cons1} \\[7pt]
%	& |a_{y}(t_k)| \leq  |a_{y_{max}}| \label{eq:FOP_cons3} \\[7pt]
%	& \vert e_{lat}(t_k)\vert \leq  l_w/2  \label{eq:FOP_cons2} \\[7pt]
%	& k \in\, [1,..., N_k]\notag
%	\end{align}
%\end{subequations}
\begin{subequations}\label{eq:FullOptProblem}
	\begin{align} 
	\min_{\bm m} \quad  [\Bar{E}_{lat}(\bm m), &\Bar{E}_{jerk}(\bm m), \Bar{E}_{v}(\bm m)] \label{eq:FOP}\\[7pt]%\]  % _{\theta_e \sim \mathcal{N}}
%	s.\,t.\quad \, \, \bm m \in\, &\bm M  \notag\\[7pt]
	s.\,t.\quad \quad \quad  \, \bm m \in\, &\bm M  \notag\\[7pt]
	 a_{x_{min}} \leq & \,{a_x}(t_k) \leq a_{x_{max}}  \label{eq:FOP_cons1} \\[7pt]
	 |a_{y}(t_k)| \leq & \,|a_{y_{max}}| \label{eq:FOP_cons3} \\[7pt]
	 \vert e_{lat}(t_k)\vert \leq &\, l_w/2  \label{eq:FOP_cons2} \\[7pt]
	 \forall k \in&\, \{1,..., N_k\}\notag
	\end{align}
\end{subequations}

where $\bm  M$ denotes the closed set of parameter constraints, $l_w$ the width of the driving lane, $a_x$ the longitudinal, $a_y$ the lateral acceleration of the ego vehicle which is limited to the maximal lateral acceleration $a_{y_{max}} = \SI{0.3}{g}$.
Ensuring that no speed limits are violated is achieved through constraints in the OCP \eqref{eq:OCP} and is therefore not considered additionally in \eqref{eq:FullOptProblem}.

\begin{figure}
	\centering
	\scriptsize
	\includegraphics[width=1.0\linewidth]{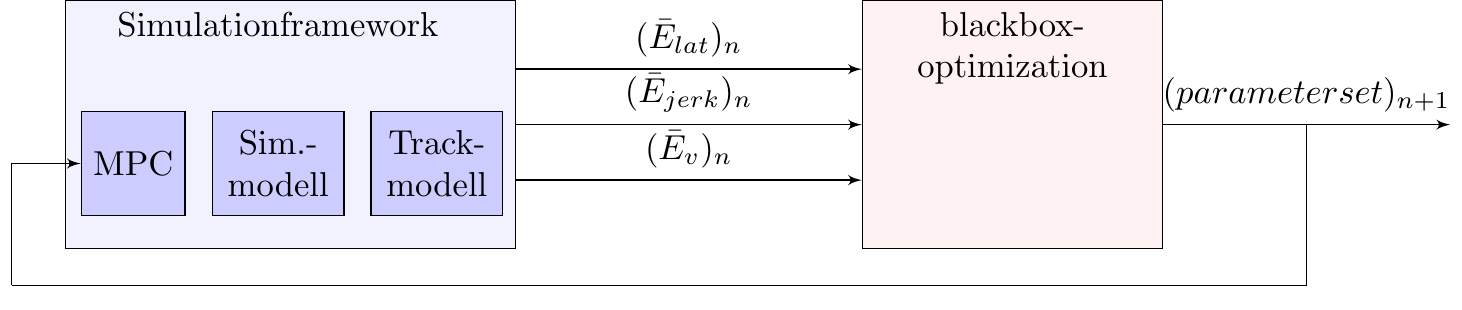}
	\caption{Overview on the Optimization environment}
	\label{fig:Opti}
\end{figure}

The aim is to obtain optimal parametrizations by maximizing comfort, dynamic, and/or path tracking accuracy. 
The workflow of the optimization and the inputs and outputs of the simulation framework are shown in Fig.~\ref{fig:Opti}. It should be noted that information about the objective function and constraints in \eqref{eq:FullOptProblem} can only be obtained by querying the simulation framework with different parametrizations. Therefore, \eqref{eq:FullOptProblem} is a black-box optimization problem.

%%%%%%%%%%%%%%%%%%%%%%%%%%%%%%%%%%%%%%%%%%%%%%%%%%%%%%%%%%%%%%%%%%%%%%%%%%%%%%%%%%

\section{Method}
\label{sec:method}

This section aims to explain how we apply Bayesian optimization (BO) to find optimal values for the weighting parameters of the MPC by solving the optimization problem stated in \eqref{eq:FullOptProblem}. 
In Section \ref{sec:BO}, Bayesian optimization is introduced. In Sections \ref{sec:BO_WS} and \ref{sec:BO_Pareto}, we will discuss two alternative approaches to solve the multi-objective optimization problem.

\subsection{Bayesian Optimization} \label{sec:BO}
%Constrained Bayesian Optimization 

%Aufbau einer BO: 

For a detailed description of Bayesian optimization, the reader is referred to \cite{Shahriari}. An overview of the numerous theoretical results such as regret bounds is also given in \cite{Shahriari}. Algorithm 1 summarizes the approach. It relies on two key elements.

The first one is a probabilistic surrogate model (cf. Step 3 of Algo. 1) which approximates all relevant Black-Box responses. In this work, Gaussian Process regression (GPR) is used. 
A separate surrogate model is created for each of the objectives $\Bar{E}_{lat}(\bm m)=O^{(1)} (\bm m), \Bar{E}_{jerk}(\bm m)=O^{(2)}(\bm m), \Bar{E}_{v}(\bm m)=O^{(3)}(\bm m)$. For simplicity they are below denoted as $O^{(i)}(\bm m)$ with $i\in \{1 , 2 , 3\}$. In addition to the objectives, a black-box response $g(\bm m)$ is modelled by GPR in order to indicate the feasibility of a parametrization. For a given evaluated parametrization $\bm m'$, $g(\bm m') = -1$, if any of the Constraints 11c - 11e were violated during its evaluation on the simulation model and $g(\bm m') = 1$ otherwise. Constraints are violated, for example if the vehicle leaves the track. 
At each iteration $n$, the GPR models are updated using all sample points obtained so far. The set $\mathcal{M}_n$ denotes all past evaluated parametrizations, whereas $\mathcal{O}_n^{(i)}$ and $\mathcal{G}_n$  denote the corresponding objective function values and observed feasibilities.   
Using these past observations, GPR is used to make normally distributed predictions for any parametrization $\bm m$:

\begin{equation}
\label{eq:probPred_y}
 	\tilde{o}^{(i)}(\bm m) \sim \mathcal{N}(\bar{o}^{(i)}(\bm m|\mathcal{M}_n,\mathcal{O}_n^{(i)}),\sigma^2_{o^{i}}(\bm m|\mathcal{M}_n,\mathcal{O}_n^{(i)}))
\end{equation}
\begin{equation}
\label{eq:probPred_g}
 	\tilde{g}(\bm m) \sim \mathcal{N}(\bar{g}(\bm m|\mathcal{M}_n,\mathcal{G}_n),\sigma^2_g(\bm m|\mathcal{M}_n,\mathcal{G}_n))
\end{equation}

Mean and standard deviation of the normally distributed predictions, $\tilde{o}$, $\tilde{g}$ are denoted by $\bar{o}$, $\bar{g}$ and $\sigma_{o}$, $\sigma_{g}$, respectively.   

The GPR model is constructed using a constant mean and a Matérn 5/2 Kernel with automated relevance detection. It is suggested by \cite{Stein1999} that clear expectations of smoothness are impractical in modeling many physical processes. Therefore they propose the class of Matérn Kernel instead of squared exponential Kernel. Hyperparameters of the Gaussian Process are optimized in each iteration by maximizing the \textit{maximum likelihood}. For a more detailed introduction to GPR, we refer to \cite{GPrasmussen}.

%For Approach 2 the hand-tuned parameterization was also selected as the initialization point.
The computational demand for the model fitting of each GPR increases cubically with the number of evaluation points.  To ensure that the computational complexity increases only approximately linearly with the  number of iterations, the sparse method Fully Independent Training Conditional (FITC) was used from 300 evaluations \cite{Shahriari}. A similar approach was chosen in \cite{stenger2019FitcSparse}.

\begin{algorithm}[h] 
	1: Initial sampling of $\mathcal{M}_{1}$, $\mathcal{O}^{(i)}_{1}$ and $\mathcal{G}_{1}$\\   [3pt]
	2: \textbf{for} n = 1; 2; . . . ; \textbf{do} \\[3pt]
	3: \quad update probabilistic  GPR surrogate models using  \\
	\hspace*{6.5mm} $\mathcal{M}_{n}$,$\mathcal{O}^{(i)}_{n}$ and $\mathcal{G}_{n} $\\[3pt]
	4: \quad select $\bm m_{n}$ by optimizing an acquisition function:\\ 
	\hspace*{6.5mm} $\bm m_{n} = \argmax_{\bm m}(\alpha(\bm m))$\\[3pt]	
	5: \quad query objective function to obtain $o^{(i)}_{n}$ and $g_{n}$ \\[3pt]
	6: \quad augment data $\mathcal{M}_{n+1} = \{ \mathcal{M}_{n},\bm m_{n}\}$, \\ \hspace*{6.5mm} $ \mathcal{O}^{(i)}_{n+1} = \{ \mathcal{O}^{(i)}_{n},o^{(i)}_{n}\}$ and
	$\mathcal{G}_{n+1} = \{ \mathcal{G}_{n},g_{n}\}$  \\[3pt]
	8: \textbf{end for} 
	\label{Algo:BayesOpt}
	\caption{Multi-objective Bayesian optimization for constrained problems}
\end{algorithm}

The second important part is the acquisition function 
\begin{equation}
\alpha(\bm m) = f(\tilde{o}^{(i)}(\bm m),\tilde{g}(\bm m)). 
\end{equation}
It assigns a scalar to each parametrization indicating how useful an evaluation of that parametrization would be for the progress of the optimization. Usually, it balances between exploration and exploitation and takes the probability of feasibility into account. It is calculated using the probabilistic predictions of the Black-Box responses $\tilde{o}^{(i)}(\bm m)$ and $\tilde{g}(\bm m)$, provided by the GPR surrogate models. Popular acquisition functions are for example \emph{Probability of Improvement}, \emph{Expected Improvement} and \emph{Entropy search}. The acquisition functions used in this contribution are explained in Sections \ref{sec:BO_WS} and \ref{sec:BO} respectively. The acquisition function is maximized using random search to obtain the next sample point $\bm m_{n}$ (cf. Step 4 of Algo. 1).

Afterward, the closed-loop behavior of $\bm m_{n}$ is evaluated using the simulation model (cf. Step 5 of Algo. 1) and the data is augmented using the obtained responses (cf. Step 6 of Algo. 1).

%In this contribution, we use the so called expected improvement with constraints \cite{EIC_constraintBO}:

%\begin{equation}
%\alpha(x) = \text{EIC}(x) = \text{EI}(\tilde{y}(x),y_{\text{min}}) \cdot %\text{P}(\tilde{g}(x) < g_{\text{max}})
%\end{equation}

%The first part balances between, sampling where the uncertainty is high (exploration) and sampling close to the current optimum (exploitation) by calculating the expected improvement (EI). The second part represents the probability of constraint fulfillment. The acquisition function is maximized using random search to obtain the next sample point $x_{n+1}$ (cf. Step 4 of Algo. 1)

%In case a certain parametrization selected by the optimization makes the car behave in a way, the simulation gets stopped, e.g. by leaving the track, this evaluation is declined as a non-feasible solution. 

\subsection{Multiobjective Optimization with weighted Sum}
\label{sec:BO_WS}

%This work will present two ways to find Pareto-optimal parametrizations for the MPC weighting parameters. - steht jetzt schon oben (David)

\begin{figure}
	\centering
	\scriptsize
	\includegraphics[width=1.0\linewidth]{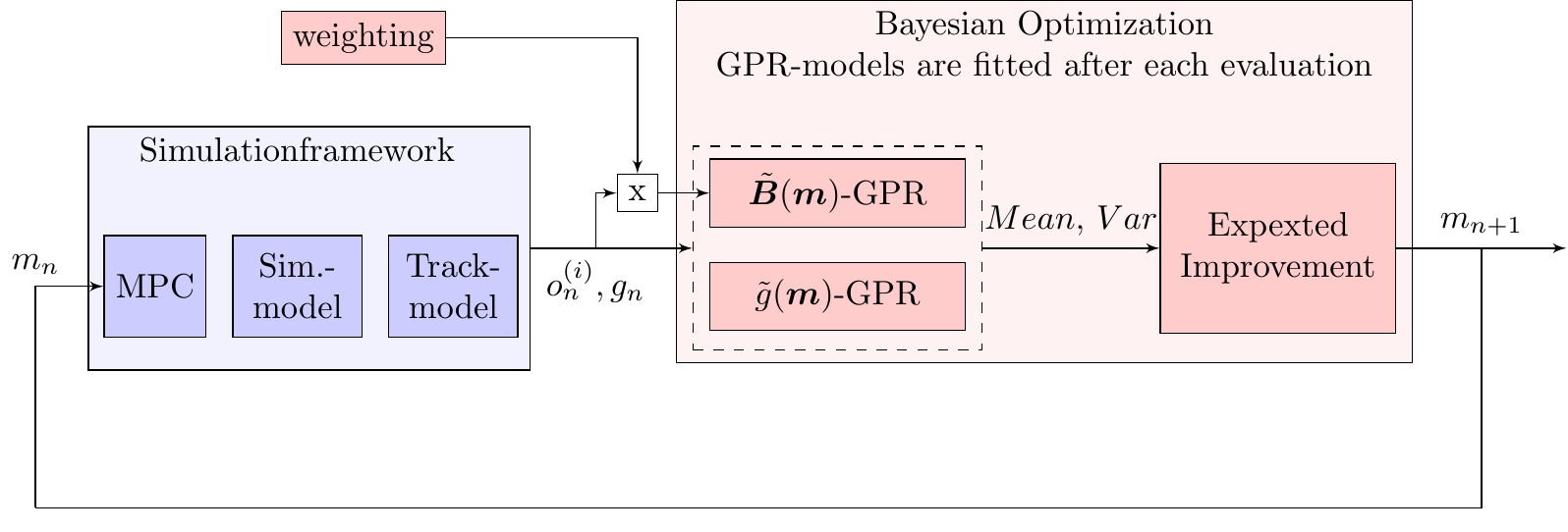}
	\caption{Optimization with weighting sum}
	\label{fig:BO_weight}
\end{figure}

%subsection – sum of scaled objectives?  
The first approach %idea 
of finding Pareto-optimal parametrizations for the MPC parameters is realized by %scaling the objectives. 
integrating the three objectives in a single objective using varying weights. Each of the differently weighted objective functions is then optimized using an instance of single-objective constraint BO. An overview of the algorithm with one instance of weighting parameters is shown in Fig.~\ref{fig:BO_weight}. 
%In total there are two GPRs that are fitted after each evaluation and seven dimensions, which will be optimized, to reach the optimum. 

The objective function $B$ is calculated by weighting the three performance criteria $\bm{\Bar{E}} = [\Bar{E}_{jerk},\, \Bar{E}_{v},\, \Bar{E}_{lat}]$ with the weights $\bm{w}= [w_1,\, w_2,\, w_3]^T$ as follows:
%%\begin{subequations}
\begin{align}
\label{eq:weightSum}
              B &= \bm{\Bar{E}} \cdot \bm{w} \notag\\ %\boldsymbol{S}_{W} &= \bm{\Bar{E}} \cdot \bm{w} \notag\\
                    &= \begin{bmatrix}\Bar{E}_{jerk}& \Bar{E}_{v}& \Bar{E}_{lat} \end{bmatrix}\cdot \begin{bmatrix}
    	        w_1\\
    	        w_2\\
    	        w_3 
    	    \end{bmatrix}.
\end{align}
Additionally, $w_1,\, w_2,\, w_3 \in [0,1]$ and $w_1 + w_2 + w_3 = 1$ hold.
%Scaling means that a certain ratio of the three criteria is optimised. 
Depending on which criterion is weighted more or less strongly, only certain areas of the Pareto-front are covered.
%To achieve different optimal solutions for different driving behavior the objectives should experience varying scaling after each optimization ({\color{red} ist verständlich, was hier gemeint ist?}). 
Thus, different weightings are cascaded (see Section \ref{sec:results}) and optimized with regard to them in order to map the entire Pareto-front. To achieve a fine resolution of the front, a fine cascading is required.

% The sum of the weighting parameters is equal to one 
% \begin{align}
%      \bm{w} = \{(w_1, w_2, w_3) \in [0,1] | w_1 + w_2 + w_3 = 1\}.
% \end{align}
% \end{subequations}
\begin{comment}
	\begin{figure}
	\centering
	%\scriptsize
	\includestandalone[width=1.0\linewidth]{texFigures/BO_weight}
	\caption{Optimization with weighting sum}
	\label{fig:BO_weight}
	\end{figure}
\end{comment}

This approach can be seen as a special case of the algorithm introduced in Section \ref{sec:BO} by using only one Objective $O^{(1)} = B$. Here the \emph{Expected Improvement with constraints}, introduced in \cite{EIC_constraintBO}, is used as the acquisition function
\begin{subequations}
\begin{align} \label{eq:EIC}
%\alpha(x) = \text{EIC}(x) = \text{EI}(\tilde{\bm{S}}_{\bm{w}}(m),\bm{S}_{\bm{w}_{\text{min}}}) \cdot \text{P}(\tilde{g}(x) < g_{\text{max}})
\alpha(\bm{m}) &= \mathrm{EIC}(\bm{m})\notag\\
&= \mathrm{EI}(\tilde{B}(\bm{m}),B_{\text{min}}) \cdot \mathrm{Pr}(\tilde{g}(\bm{m}) > g_{\text{max}})
\end{align}
with
\begin{align}
    \label{eq:EIC-EI}
    \mathrm{EI}(\tilde{B}(\bm{m}),B_{\text{min}}) &= \sigma_{B}(\bm{m})  (z \cdot \Phi(z) + \phi(z))\\
    z &= \frac{B_{\text{min} }-\bar{B}(\bm{m})}{\sigma_{B}(\bm{m}) }
\end{align}
\end{subequations}
%\intertext{The expected improvement is calculated as follows}
where $\Phi$ and $\phi$ denote the cumulative distribution function and the probability density function of the normal distribution. The best objective function value obtained so far is denoted by $B_{\text{min}}$ and the probabilistic predictions for the objective function and constraints are denoted by $\tilde{B}(\bm{m})$ and $\tilde{g}(\bm{m})$ (cf. Eq. \eqref{eq:probPred_y} and Eq. \eqref{eq:probPred_g}). 
The Expected improvement $\mathrm{EI}(\tilde{B}(\bm{m}),B_{\text{min}})$ balances between exploitation and exploration by assigning a scalar utility value to each parametrization $\bm{m}$. It increases, if the predicted uncertainty of the objective function increases (exploration) or the predicted mean of the objective function decreases (exploitation). The second part of Eq. \eqref{eq:EIC} represents the probability of constraint fulfillment with $g_{\text{max}} = 0$.

\subsection{Pareto Optimization}
\label{sec:BO_Pareto}

% hier evtl. etwas zu Pareto-Optimalen Punkten erzählen? 
% Pareto-optimal points are non-dominant in all criterions. For a pareto-optimal point you can’t find any other point, which is better (dominant) in all criterions.

%subsection –Multiobjective optimization?  
%An alternative way to calculate non-dominated points is shown in Fig.~\ref{fig:BO_Pareto}. 
\begin{comment}
\begin{figure}
	\centering
	%\scriptsize
	\includestandalone[width=1.0\linewidth]{texFigures/BO_Pareto}
	\caption{Pareto Optimization}
	\label{fig:BO_Pareto}
\end{figure}
\end{comment}
\begin{figure}
	\centering
	%\scriptsize
	\includegraphics[width=1.0\linewidth]{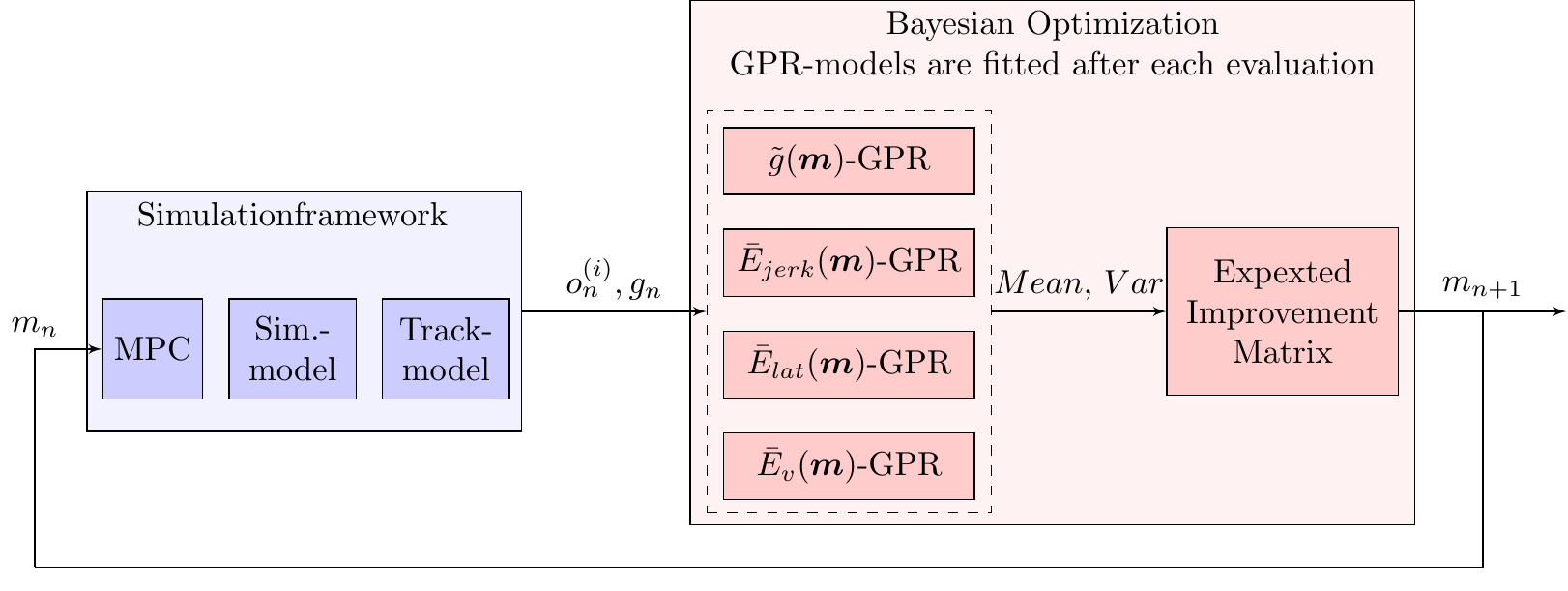}
	\caption{Pareto Optimization}
	\label{fig:BO_Pareto}
\end{figure}

The idea of the second method is based on Pareto optimization (see Fig.~\ref{fig:BO_Pareto}). This means that all three criteria are optimised simultaneously in a single instance of Bayesian optimization without the necessity of cascaded weighting of the objectives. In contrast to \textit{standard} BO, a Pareto-front of non-dominant parametrizations representing possible compromises is searched for.
Therefore, each objective is modeled by a separate instance of GPR as described in Sec. \ref{sec:BO}. In combination with the feasibility-GPR, the total number of GPRs is four. %The standard  Expected improvement acquisition function is not suited of pareto optimization. 

The \emph{Expected Improvement Matrix Criteria (EIM)} presented in \cite{EIM_Zhan2017ExpectedIM} extend the well-known single-objective EI to the multi-objective case. Here the euclidean distance-based $\mathrm{EIM_{e}}$ criterion is used in combination with the probability of feasibility as the acquisition function: %which results in a single optimization to reach a Pareto-front of non-dominant points. 
%The acquisition function  
\begin{align}
&\alpha(\bm{m})   = \mathrm{CEIM}(\bm{m}) = \\
&\quad   \mathrm{EIM_{e}}(	\tilde{o}^{(1)}(\bm m),...,\tilde{o}^{(3)}(\bm m),\mathcal{O}_{\text{min}}) \cdot \mathrm{Pr}(\tilde{g}(\bm{m}) < g_{\text{max}}) \notag
\end{align} 
%will be maximized to decide which next point will be queried, by trading off exploration and exploitation, to achieve efficient sampling. 
 
The set of Pareto-optimal points obtained so far is denoted by $\mathcal{O}_{\text{min}}$. The $\mathrm{EIM_{e}}$ criterion seeks to extend and further detail the current pareto front. However, the size of the Expected Improvement Matrix increases linearly with the number of pareto-optimal points in $\mathcal{O}_{\text{min}}$ and with the number of objectives. Therefore the calculation time also increases with the number of pareto-optimal points. Additional details regarding the computational cost can be found in \cite{EIM_Zhan2017ExpectedIM}.    

%Since the update of the GPR model is the more computationally intensive part of the BO and increases cubically with increasing evaluation points, it is expected that with this algorithm the computing time increases significantly more with increasing evaluation numbers than the algorithm with the weighted sum. %\textcolor{red}{hier noch was zu sparse schreiben, oder soll das bei den sowie bisher gehabt bei den Results bleiben? --> kurz + Quelle}

\section{Results}
\label{sec:results}
The algorithms presented in Sections \ref{sec:BO_WS} and \ref{sec:BO_Pareto} to solve the optimization problem \eqref{eq:FullOptProblem} were implemented in Matlab 2018b using the GPML toolbox \cite{GPML} to create GPR models. 
In Section \ref{sec:results_A}, results of the approach with weighted objectives will be presented and compared to a parametrization, which was hand-tuned by an expert. 
Since the two algorithms
\begin{itemize}
    \item Approach 1: Optimization with weighted sum  (cf. Section \ref{sec:BO_WS})
    \item Approach 2: Pareto optimization (cf. Section \ref{sec:BO_Pareto})
\end{itemize}
were able to achieve similar Pareto-fronts, only achieved parameterisations of Approach 1 are presented.
%Three parameter sets, which are part of the Pareto front will be chosen and on the base of the vehicle behavior assigned to a driving scenario, it fits to. 
Afterward, the performance of the two algorithms will be compared.

\subsection{Comparison of Pareto-optimal Parametrizations to hand-tuned Parametrization}
\label{sec:results_A}
% Vergleich zwischen 3 Paretopunkten mit der Ursprungsparametrierung

%First of all the car's behavior with the hand-tuned parameterization is compared with three selected Pareto points.  

\begin{comment}
\begin{figure}
\centering
\scriptsize
%	\includestandalone[width=1.0\linewidth]{texFigures/Pareto_marked}	
\includestandalone[width=1.0\linewidth]{texFigures/Pareto_marked_new}
\caption{Pareto-optimal points achieved by Approach 1 and the point belonging to the hand-tuned parametrization. The three markers (1 to 3) indicate Pareto-optimal points discussed in the text.}
\label{fig:Pareto_marked}
\end{figure}
\end{comment}
\begin{figure}
	\centering
	\scriptsize
	\includegraphics[width=1.0\linewidth]{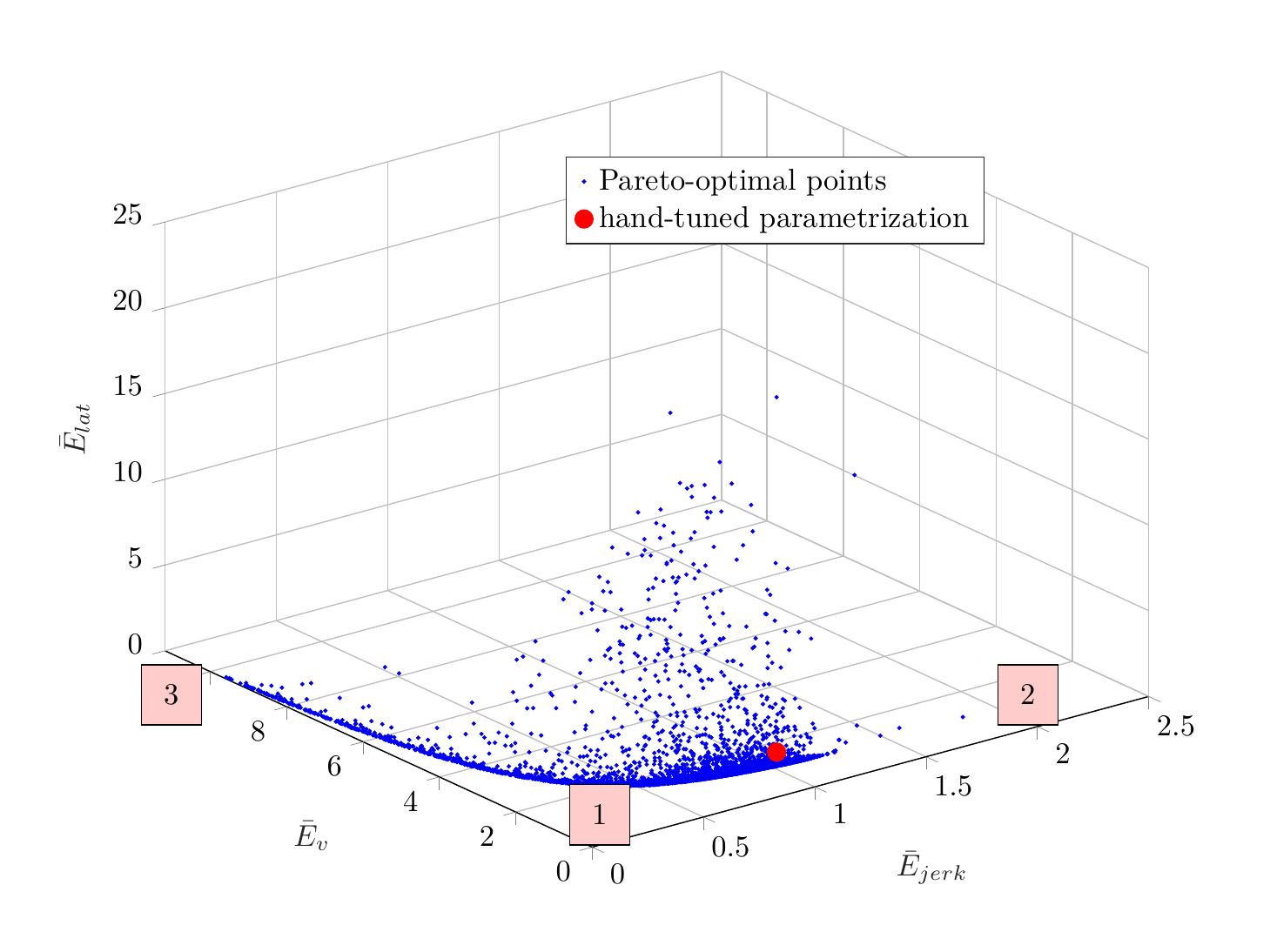}
	\caption{Pareto-optimal points achieved by Approach 1 and the point belonging to the hand-tuned parametrization. The three markers (1 to 3) indicate Pareto-optimal points discussed in the text.}
	\label{fig:Pareto_marked}
\end{figure}

Fig.~\ref{fig:Pareto_marked} shows the Pareto-optimal set obtained by Approach 1. The objective function values are scaled to the results from the hand-tuned expert parametrization. Therefore, the expert parametrization yields objective function values of $(1\, 1\, 1)$, see red dot in Fig.~\ref{fig:Pareto_marked}. It can be seen that a rich set of Pareto-optimal parametrizations is found. This enables control engineers to choose from a number of competitive options depending on the desired driving behavior.  

%The objectives, that are shown in Fig.~\ref{fig:Pareto_marked} are scaled to the results from the default parametrization, which means, that the hand-tuned parametrization's objectives are all equal to (1 1 1). 

\begin{figure}
	\centering
	\scriptsize
	\includegraphics[width=1.0\linewidth]{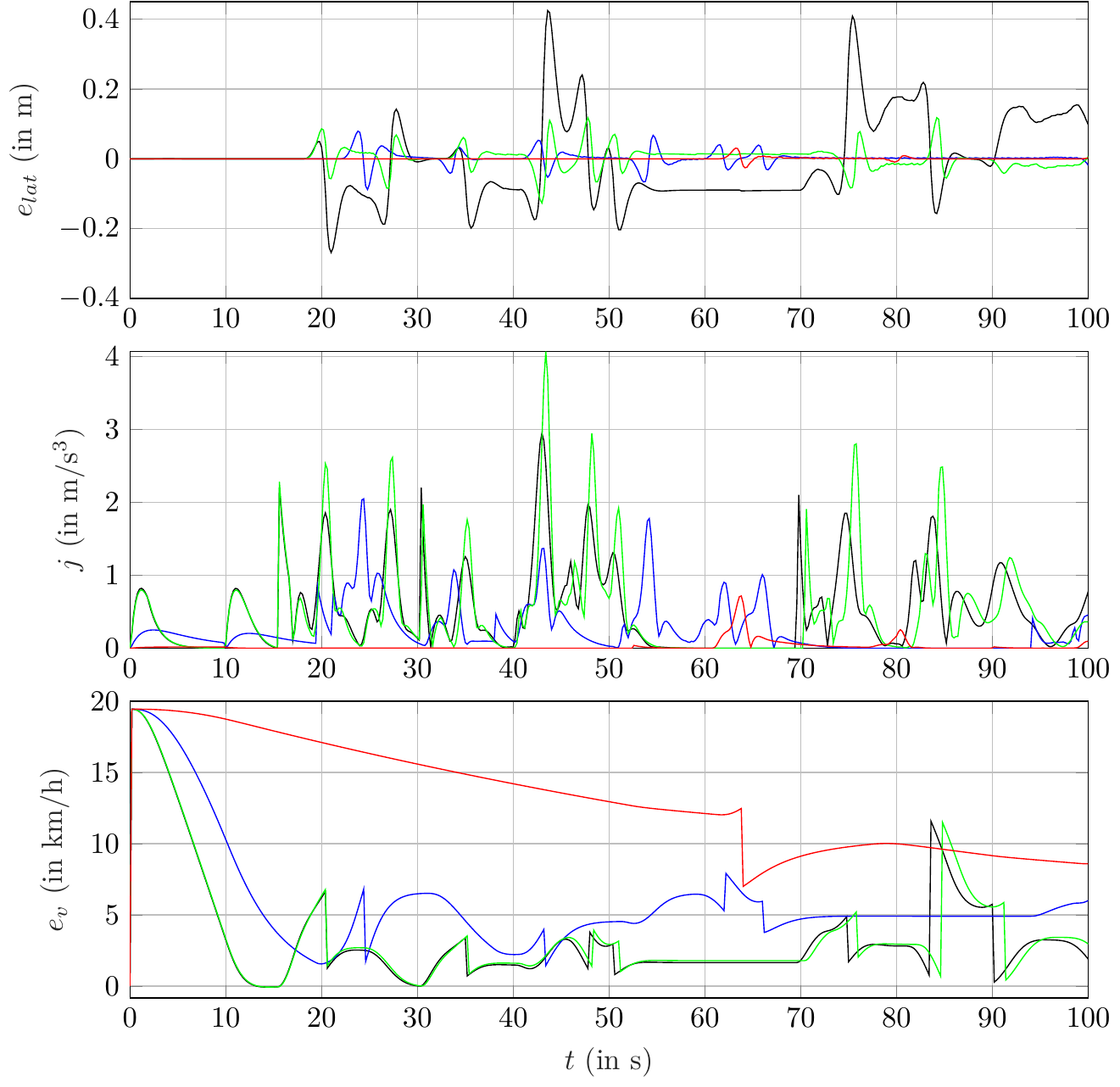}
	\caption{Lateral deviation $e_{lat}$, jerk $j$ and velocity error $e_v$ of the ego vehicle belonging to the three selected Pareto-optimal points marked in Fig.~\ref{fig:Pareto_marked} (1~=~blue, 2 = green, 3 = red) in comparison with the hand-tuned parametrization (black)}
	\label{fig:3CompPointsToDefault}
\end{figure}

%Further analysis will be based on Fig. \ref{fig:3CompPointsToDefault}, which shows the lateral deviation, the jerk and velocity error vehicle signals with the hand-tuned parametrization and the in Fig.~\ref{fig:Pareto_marked} three marked Pareto-optimal points. 
Fig.~\ref{fig:3CompPointsToDefault} shows the lateral deviation $e_{lat}$, the jerk $j$ and the velocity error $e_v$ of the ego vehicle for the hand-tuned parametrization and for the parametrization belonging to the three selected Pareto-optimal points marked in Fig.~\ref{fig:Pareto_marked}. Note that only the first 100 seconds are presented for better clarity.

 It can be seen, that the hand-tuned parametrization is almost pareto-optimal. However, in comparison to point 2, it is characterized by a slightly lower velocity error at the cost of substantially higher lateral deviation (cf Fig. \ref{fig:3CompPointsToDefault} top). Therefore, the hand-tuned parametrization is more suitable for dynamical driving requirements.

\begin{comment}
\begin{figure}
\centering
\scriptsize
\includestandalone[width=1.0\linewidth]{texFigures/3CompPointsToDefault}
\caption{Lateral deviation $e_{lat}$, jerk $j$ and velocity error $e_v$ of the ego vehicle belonging to the three selected Pareto-optimal points marked in Fig.~\ref{fig:Pareto_marked} (1~=~blue, 2 = green, 3 = red) in comparison with the hand-tuned parametrization (black)}
\label{fig:3CompPointsToDefault}
\end{figure}
\end{comment}

%Point 2 is an example of a parameterization with which the vehicle behaves only slightly slower than with the hand-tuned parameterization. However, the vehicle with the parameterization of point 2 has a considerably lower track deviation.
With the parameterization of point 3, the vehicle drives very slowly, but very comfortably and with a very high level of accuracy in tracking. Due to the good lateral tracking and low jerk, the velocity tracking error is rather high.

The parameterization of point 1 can be seen as a reasonable trade-off between 2 and 3. %In comparison, the vehicle drives slightly more comfortable than the hand-tuned parametrization, because jerk and lateral deviation are smaller, but the vehicle behaves slower because the lap time on the circuit is higher and the value for the speed deviation from the limit is higher.
In comparison to the hand-tuned parametrization, the vehicle drives slightly more comfortably, which can be recognized by the fact that jerk and lateral deviation are smaller. However, the vehicle behaves less agile, which can be seen by the higher lap time for the track and the higher velocity deviation from the limit.
Because high tracking and low jerk are characteristics for comfortable driving, this parameterization is suitable for highway driving, since high-speed changes are not necessary.

\subsection{Comparison of Pareto Optimization and Optimization with weighted Sum}

In the following section the two Approaches 1 and 2 are compared with each other, regarding their number of evaluations, run time and their performance. %Subsequently, the results achieved by the two algorithms are compared with each other. 
For the calculation of the performance indicator, a script of the PlatEMO \cite{PlatEMO} platform was used, which calculates the hypervolume indicator (HV-indicator) according to Zitzler et al \cite{ZitzlerThiele_MultiObj}.

%\begin{itemize}
%    \item Algo. 1: Optimization with weighted sum  (cf. Section \ref{sec:BO_WS})
%    \item Algo. 2: Pareto-optimization (cf. Section \ref{sec:BO_Pareto})
%\end{itemize}

 \begin{table}[ht] %Resultstable
	\caption{Performance characteristics of both approaches}
	\centering
	\begin{tabular}{ l r r r r  }
		\toprule
		Approach:           &   1          & 2   \\
		\midrule
		Evaluations          &   19800   &  2500 \\ % +-0.6  
		Run time (in s)      &   6.45e+05 	  & 7.81e+05 \\ %\textcolor{red}{8.07e+05}
		HV-Indicator         &   0.8517    &  0.8480      \\
		%\hline
		\bottomrule
	\end{tabular}
	\label{tab:Results1D}
\end{table}

Table~\ref{tab:Results1D} shows the performance achieved at the end of both algorithms.
For Approach 1, 66 different weightings were chosen with 300 evaluation points each. This is to ensure that the parameters of the MPC optimize the driving behavior concerning all three criteria/goals. Preliminary experiments have shown, that the optimizer has (nearly) converged within 300 evaluations. This leads to 19800 points, that were evaluated for Approach 1.% The total calculation time is 6.45e+05 seconds.
%As a comparison, for Algo. 2 2500 iterations were performed.
For Approach 2 the hand-tuned parameterization was also selected as the initialization point. %The calculation time here was 7.81e+05 seconds. 
%Since the computing capacity for each GPR increases cubically with the number of evaluation points (model fitting), the computing time for Approach 2, which contains four GPRs (three GPRs for the targets and one for the feasibility GPR), would increase $4n^3$. To ensure that the computational capacity for a GPR increases only approximately linearly with the number of iterations, the sparse method (FITC) was used from 300 evaluations on \cite{stenger2019FitcSparse}.

\begin{figure}
	\centering
	\scriptsize
	\includegraphics[width=1.0\linewidth]{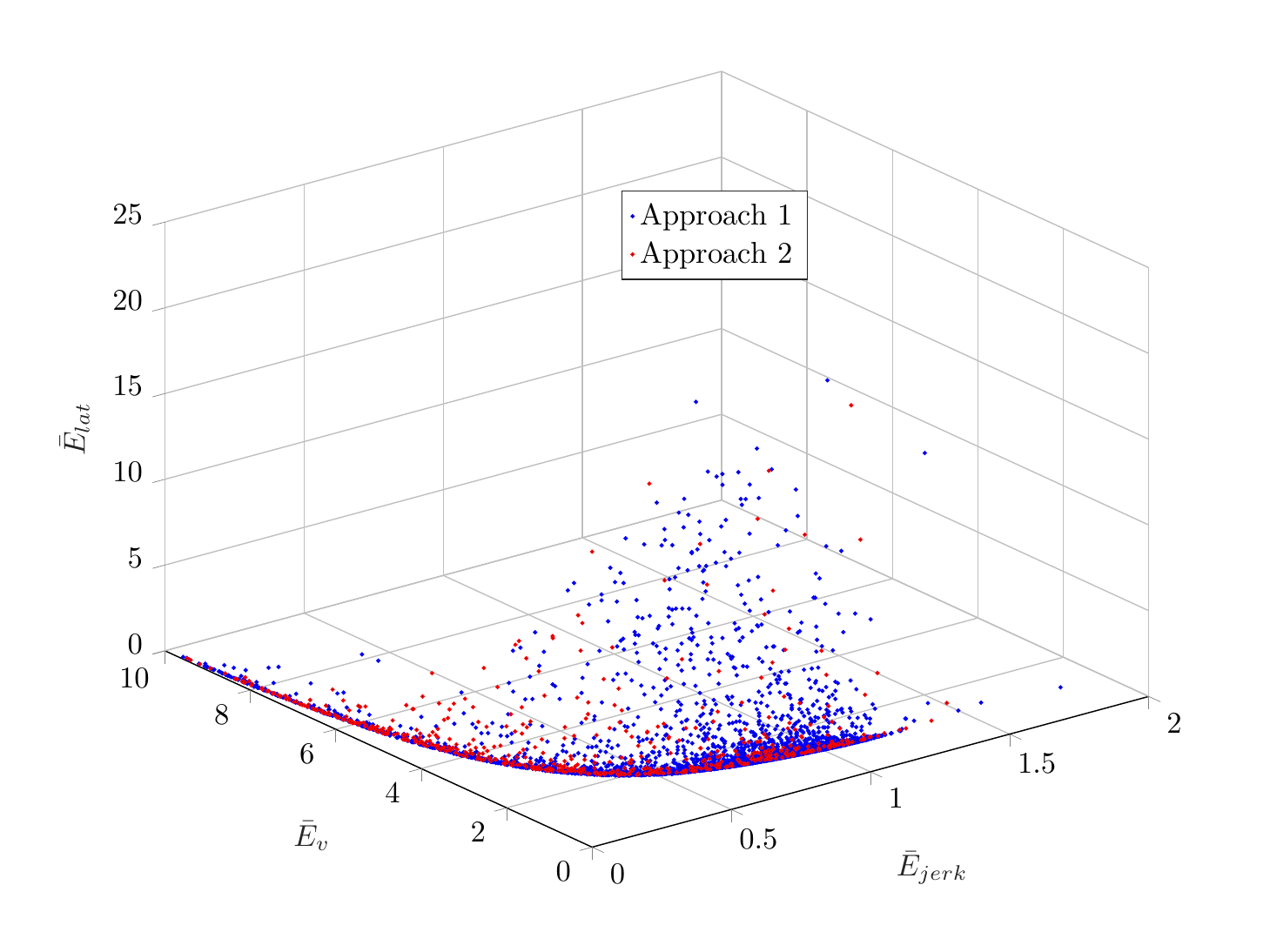}
	\caption{Pareto-optimal evaluations of both approaches}
	\label{fig:Plot3Dcomp}
\end{figure}

% Vergleich der beiden Algorithmen mithilfe von (ETH-Paper zitieren und PlatEMO?) und Abbildung
%Subsequently, the results achieved by the two algorithms are compared with each other. For the calculation of the performance indicator, which is based on a calculation of the hypervolume indicator (HV-indicator) according to Zitzler et al \cite{ZitzlerThiele_MultiObj}, a script of the PlatEMO \cite{PlatEMO} platform was used.
% ------------------------------- Reference --------------------------------
% E. Zitzler and L. Thiele, Multiobjective evolutionary algorithms: A
% comparative case study and the strength Pareto approach, IEEE
% Transactions on Evolutionary Computation, 1999, 3(4): 257-271.
  
%The two algorithms have achieved the following performance:  
%\begin{itemize}
%    \item Algo. 1: 0.8517 %\textcolor{red}{0.8796}
%    \item Algo. 2: 0.8480 %\textcolor{red}{0.8765}
%\end{itemize}

Fig.~\ref{fig:Plot3Dcomp} shows all non-dominant points of the two algorithms. It can be seen that Approach 1 achieves a finer resolution of the Pareto-front than Approach 2 in area of high jerk (large $\bar{E}_{jerk}$). 
\begin{comment}
\begin{figure}[ht]
\centering
\includestandalone[width=1\linewidth]{texFigures/subplots_test}
\caption{Top: Comparison of the HV-indicator as a function of accumulated computing time. Bottom: Accumulated computing time as a function of evaluations for Approach 1 (blue) and Approach 2 (red)}
\label{fig:comp_subplots}
\end{figure}
\end{comment}
\begin{figure}[ht]
	\centering
	\includegraphics[width=1\linewidth]{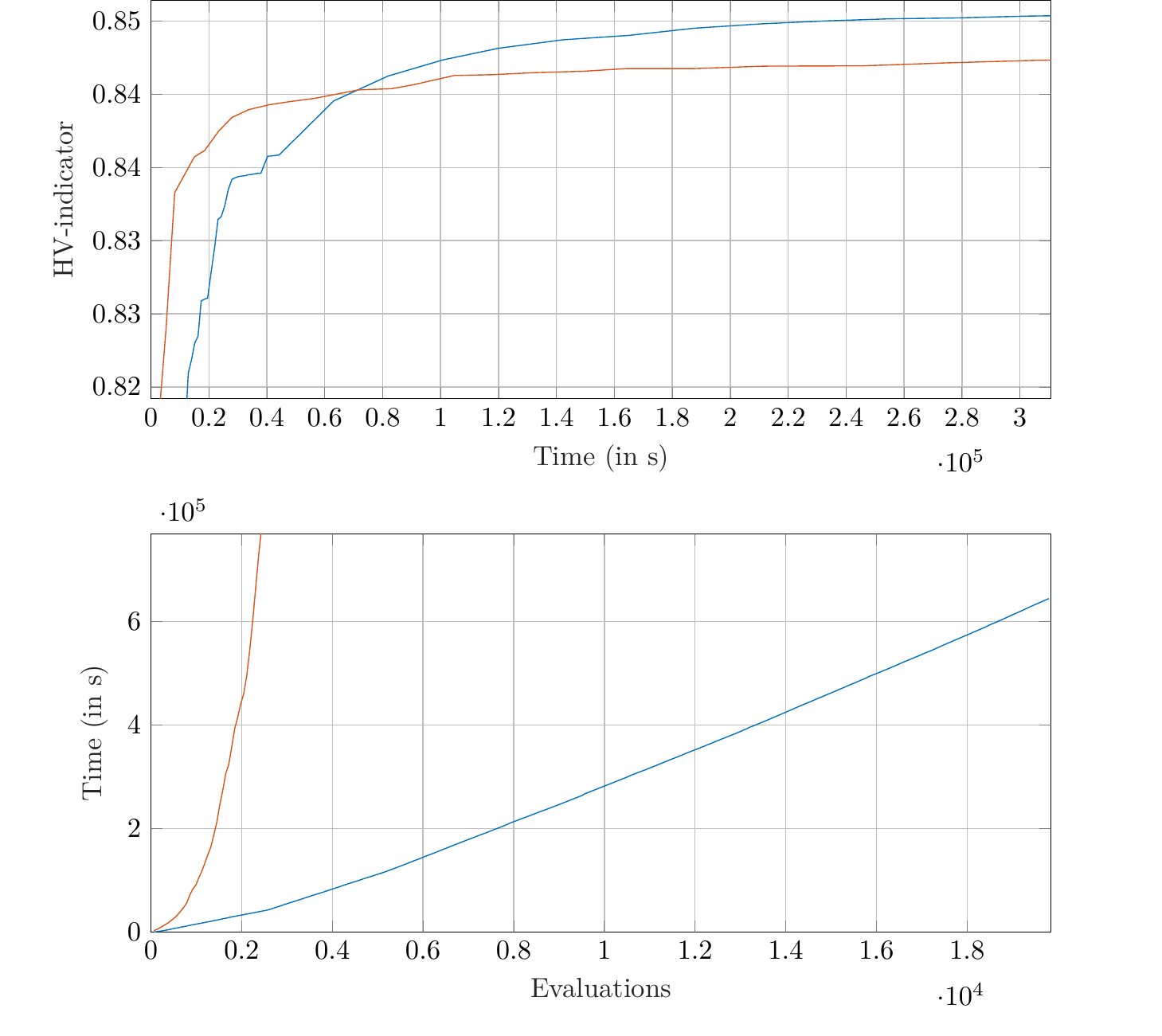}
	\caption{Top: Comparison of the HV-indicator as a function of accumulated computing time. Bottom: Accumulated computing time as a function of evaluations for Approach 1 (blue) and Approach 2 (red)}
	\label{fig:comp_subplots}
\end{figure}

At the end of the optimization methods, Approach 1 achieves better performance in terms of the HV-indicator with more total evaluations despite the lower computing time. However, Fig. \ref{fig:comp_subplots} shows that for low computing times Approach 2 is superior to Approach 1 with regard to the HV-indicator. 

%%%%%%%%%%%%%%%%%%%%%%%%%%%

%\addtolength{\textheight}{-5cm}
\addtolength{\textheight}{-1cm}

%%%%%%%%%%%%%%%%%%%%%%%%%%%%%%%%%%%%%%%%%%%%%%%%%%%%%%%%%%%%%%%%%%%%%%%%%%%%%%%%

Additionally, it can be seen that the cumulative computing time increases only approximately linearly with the number of evaluations for Approach 1. This means that the overhead generated by BO is almost constant, which can be attributed to the optimization being reset completely after each of the 66 weight-combinations. Therefore, at most 300 evaluations are used to fit the GPR model. In contrast, the overhead per iteration of Approach 2 increases substantially during optimization. This can be attributed to two aspects of the algorithm. First, unlike as in Approach 1, all points evaluated so far are used to find the next sample point resulting in an increasing computing time for the 4 Gaussian Process models. Second, the calculation time of the acquisition function increases with the growing number of points in the pareto front.
In summary, for the presented case, Approach 2 is more sample efficient initially, but less scalable than Approach 1.

\section{Conclusion}
In this paper, two multi-objective Bayesian optimization approaches for simulation-based parameter tuning of a model predictive path-following control realization for autonomous driving have been presented. As optimization objectives comfort, dynamic and path tracking accuracy are used, which are described by defined characteristic values. The result is a Pareto-front, which can be used to select MPC parameter sets for different applications. The obtained results show the effectiveness of the proposed approaches to find Pareto-optimal parametrizations for an MPC for vehicle guidance. Furthermore, the comparison of both approaches indicated that both approaches achieve similar results, but differ in the required evaluations and computing time as well as the convergence rate.

%%%%%%%%%%%%%%%%%%%%%%%%%%%%%%%%%%%%%%%%%%%%%%%%%%%%%%%%%%%%%%%%%%%%%%%%%%%%%%%%
%\section*{APPENDIX}


\section*{ACKNOWLEDGMENT}
We gratefully acknowledge the financial support of this work by the  Federal Ministry for Economic Affairs and Energy under the grant 50NA1912.

%%%%%%%%%%%%%%%%%%%%%%%%%%%%%%%%%%%%%%%%%%%%%%%%%%%%%%%%%%%%%%%%%%%%%%%%%%%%%%%%

\bibliographystyle{IEEEtran}
\bibliography{ms}

\end{document}